\documentclass[letterpaper, 10 pt, journal, twoside]{IEEEtran}
\IEEEoverridecommandlockouts
\usepackage[ruled,vlined,linesnumbered]{algorithm2e}
\usepackage{cite}
\usepackage{amsmath,amssymb,amsfonts,algorithmic,textcomp, multirow}
\usepackage{graphicx,subfig,float,xcolor,url,hyperref,etoolbox}
\usepackage[nolist]{acronym}
\hypersetup{
    colorlinks,
    linkcolor={blue!50!black},
    citecolor={blue!50!black},
    urlcolor={black!80!black}
}

\DeclareMathOperator*{\argmax}{arg\,max}

\usepackage{tabularx}
\usepackage{soul}
\usepackage{adjustbox}
\usepackage{svg}

\def\BibTeX{{\rm B\kern-.05em{\sc i\kern-.025em b}\kern-.08em
    T\kern-.1667em\lower.7ex\hbox{E}\kern-.125emX}}

\makeatletter
  \patchcmd{\@maketitle}
   {\addvspace{0.5\baselineskip}\egroup}
   {\addvspace{-0.1\baselineskip}\egroup}
   {}
   {}
\makeatother
\SetKw{KwBy}{by}
\SetKw{KwNot}{not}
\makeatletter
\newcommand\notsotiny{\@setfontsize\notsotiny\@vipt\@viipt}
\makeatother

\newcommand{\CSafety}{C_{S}}
\newcommand{\CReachability}{C_{R}}
\newcommand{\CAffordance}{C_{A}}
\newcommand{\ObjectPose}{\Psi_O }
\newcommand{\HHandPose}{\Psi_{hh}}
\newcommand{\HFacePose}{\Psi_{hf}}
\newcommand{\RobotEE}{\Psi_{ree}}

\newcommand{\RobotGrasp}{g_r}
\newcommand{\HumanGrasp}{g_h}

\begin{document}

\title{\LARGE Affordance-Aware Handovers with Human Arm Mobility Constraints}

\author{Paola Ard{\'o}n$^{*\dag}$, Maria E. Cabrera$^{\dag}$, {\`E}ric Pairet$^{*}$, Ronald P. A. Petrick$^{*}$, Subramanian Ramamoorthy$^{*}$,\\ Katrin S. Lohan$^{*}$ and Maya Cakmak$^{\dag}$ \vspace{-0.2cm}
\thanks{Manuscript received: October 14, 2020; Revised January 4, 2021; Accepted February 7, 2021. This paper was recommended for publication by Editor Dan Popa upon evaluation of the Associate Editor and Reviewers' comments.}
\thanks{This research was done while the first author was on an academic visit to the University of Washington. It is supported by the Scottish Informatics and Computer Science Alliance (SICSA), EPSRC ORCA Hub (EP/R026173/1) and consortium partners. $^{*}$Edinburgh Centre for Robotics at the University of Edinburgh and Heriot-Watt University, Edinburgh, Scotland, UK. $^{\dag}$Paul G. Allen School of Computer Science \& Engineering, University of Washington, Washington, USA.
\tt paola.ardon@ed.ac.uk}
}

\begin{acronym}[ransac]
  \acro{LbD}{learning by demonstration}
  \acro{RL}{reinforcement learning}
  \acro{SVM}{Support Vector Machine}
  \acro{DoF}{degrees-of-freedom}
  \acro{CAD}{computer-aided design}
  \acro{ROI}{regions of interest}
  \acro{MRF}{Markov random fields}
  \acro{ECV}{early cognitive vision}
  \acro{IADL}{instrumental activities of daily living}
  \acro{CDR}{cognitive developmental robotics}
  \acro{2-D}{two-dimensional}
  \acro{3-D}{three-dimensional}
  \acro{RANSAC}{random sample consensus}
  \acro{RGB-D}{red-green-blue depth}
  \acro{IFR}{International Federation of Robotics}
  \acro{CNN}{convolutional neural network}
  \acro{KB}{knowledge base}
  \acro{MSE}{mean square error}
  \acro{MLN}{Markov logic network}
  \acro{XAI}{explainable artificial intelligence}
  \acro{MC-SAT}{model-constructing satisfiability calculus}
  \acro{WCSP}{weighted constraint satisfaction problem}
  \acro{MAP}{Maximum--Likelihood}
  \acro{O-CNN}{octree-based convolutional neural networks}
  \acro{OACs}{object-action complexes}
  \acro{CAD}{computer-aided-design}
  \acro{ROC}{receiver operating characteristics}
  \acro{AUC}{area under the curve}
  \acro{MCMC}{Markov chain Monte Carlo}
  \acro{FOL}{first-order logic}
  \acro{DMP}{dynamic movement primitive}
  \acro{M-RCNN}{mask RCNN}
  \acro{SAGAT}{self-assessment of grasp affordance transfer}
  \acro{OTP}{object transfer point}
  \acro{CHSS}{Chest Heart and Stroke Scotland}
  \acro{ALS}{amyotrophic lateral sclerosis}
  \acro{SRL}{statistical relational learner}
  \acro{WCSP}{weighted constraint satisfaction problem}
  \acro{ANOVA}{analysis of variance}
  \acro{ARAT}{action research arm test}
  \acro{ADL}{activities of daily living}
\end{acronym}

\markboth{IEEE Robotics and Automation Letters. Preprint Version. Accepted February, 2021}{Ard\'on \MakeLowercase{\textit{et al.}}: Affordance-Aware Handovers}
\maketitle

\begin{abstract}
Reasoning about object handover configurations allows an assistive agent to estimate the appropriateness of handover for a receiver with different arm mobility capacities. While there are existing approaches for estimating the effectiveness of handovers, their findings are limited to users without arm mobility impairments and to specific objects. Therefore, current state-of-the-art approaches are unable to hand over novel objects to receivers with different arm mobility capacities. We propose a method that generalises handover behaviours to previously unseen objects, subject to the constraint of a user's arm mobility levels and the task context. We propose a heuristic-guided hierarchically optimised cost whose optimisation adapts object configurations for receivers with low arm mobility. This also ensures that the robot grasps consider the context of the user's upcoming task, i.e., the usage of the object. To understand preferences over handover configurations, we report on the findings of an online study, wherein we presented different handover methods, including ours, to $259$ users with different levels of arm mobility. We find that people's preferences over handover methods are correlated to their arm mobility capacities. We encapsulate these preferences in a \ac{SRL} that is able to reason about the most suitable handover configuration given a receiver's arm mobility and upcoming task. Using our \ac{SRL} model, we obtained an average handover accuracy of $90.8\%$ when generalising handovers to novel objects.

\end{abstract}

\section{Introduction}\label{sec:intro}

    \IEEEPARstart{M}{any} scenarios in which robots assist humans inevitably involve robot-to-human {\em object handovers}---the transfer of objects from a robot to a human \cite{ortenzi2020object}. Successful handovers are an essential part of tasks in different domains, from fetching medication for older adults in their home, to handing tools to workers in a factory. 
    Beyond successfully transferring objects, handovers should minimise  effort needed from the human. This not only includes effort to {\em take} the object, but also effort to {\em use} the object afterwards. For example, imagine a robot handing over a bottle to a person who intends to drink from it. The robot’s choice of how to grasp and locate the bottle for the exchange determines how the person will take the object. Hence, in making those choices, the robot should aim to minimise the human's need to extend their arm, offering the bottle in a pose that facilitates drinking without needing to re-grasp the bottle. A method able to adapt robot handovers, with the goal of minimising the person's effort, is particularly convenient for users with arm mobility impairments, where usually the mobility condition changes over time \cite{sampson2015using}.
    
    Prior work models the human as able-bodied, with no mobility limitations in reaching for and grasping the object presented by the robot. These works learn human preferences over object poses~\cite{cakmak2011human}, transfer locations~\cite{sisbot2012human,nemlekar2019object}, and limited grasp locations on selected objects based on their use (i.e., object affordance) \cite{chan2014determining,bestick2016implicitly} that are inaccessible to people with mobility impairments. As such, the current literature is not able to generalise robot grasps and object poses to users with different arm mobility capacities. In this paper we aim to address this gap by explicitly considering the range of mobility constraints that the human receiver might have. 
    
    \begin{figure}[t!]
      \centering
      \includegraphics[width= 8.5cm]{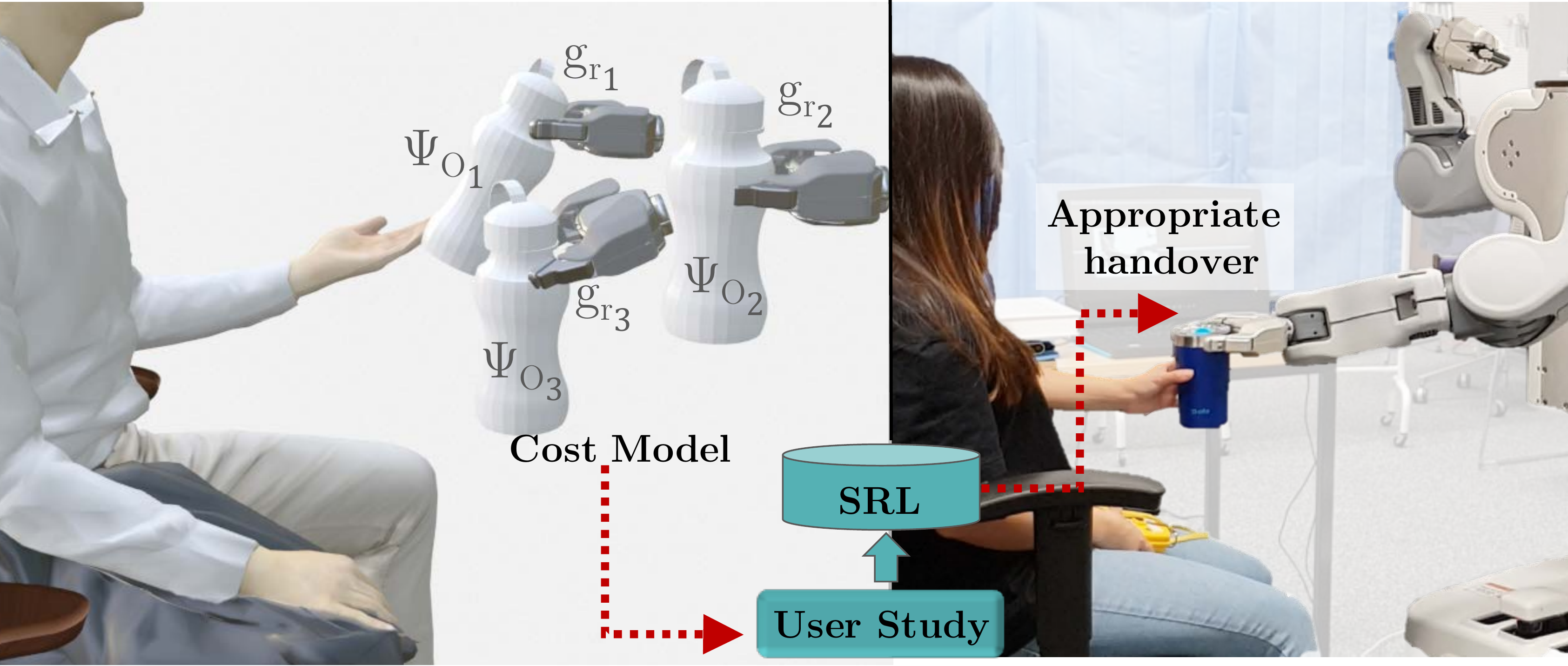}
      \caption{On the left, simulated generation of robot grasps $\RobotGrasp$ and object poses $\ObjectPose$ for handovers using our proposed cost model. On the right, real world deployment of a found suitable handover using our learned \ac{SRL} model, given the user arm mobility and upcoming task. 
      \label{fig:intro}}
      \vspace{-0.3cm}
    \end{figure}
    We present a method for automatically selecting handover grasps and poses by explicitly taking into account differences in the human receiver's arm mobility while minimising effort. We consider the handover to be composed of a suitable robot grasp that considers the receiver’s
    upcoming task, and an object pose that is safe and reachable depending on the user’s arm mobility level. A summary of our approach is depicted in Fig.~\ref{fig:intro}. Firstly, we pose the problem as hierarchical optimisation with a cost model that adapts to the receiver's mobility constraints, while considering the intended use of the object. Secondly, we evaluate our model through an online survey in which $259$ participants with mixed arm mobility limitations rate different handover poses, including the ones generated with our method. An analysis of the responses shows that handover preferences vary significantly across users with different arm mobility capacities, with mobility impaired individuals showing higher preference towards handovers selected with our method. Finally, we extend our method to generate handover configurations for previously unseen objects using a \acf{SRL}. Experimental evaluation of the \ac{SRL} handover model demonstrates generalisation of affordance-aware handovers, obtaining an average handover pose accuracy of $90.8\%$ across different mobility levels and upcoming tasks with novel objects.
    
\section{Related Work}\label{sec:related_work}
    
    Robotic handover research has taken on increasing importance due to numerous use cases in industrial and domestic assistance scenarios \cite{ortenzi2020object}. Usually, robot-to-human and human-to-robot handovers are studied separately, due to the differing nature of the challenges involved. As our focus is on robot-to-human handovers in home-assistance setups, we limit our discussion of related work accordingly. 
    
    Robot-to-human handovers have broadly focused on learning preferences over how the object should be transferred. Some of the factors that have been considered are the effect of gaze \cite{admoni2014deliberate,moon2014meet}, different object poses \cite{cakmak2011human}, suitable distance from robot to human \cite{nemlekar2019object,shi2013model}, and optimal duration \cite{huang2015adaptive, huber2008human, tang2020assessment} for the handover task. Few works include in their object transfer policy the extraction of suitable robot grasps considering the receiver's upcoming task \cite{chan2014determining,bestick2016implicitly}. 
    
    In general, the current literature extracts preferences, over the previously mentioned factors, through user studies where the participants do not report arm mobility impairments. Consequently, the state-of-the-art handover models are not inclusive and cannot be mapped to users with different arm mobility capacities. In contrast, our approach generalises across users with different arm mobility levels while lowering the receiver's effort during and after the handover task. We consider (i)~the object affordances to extract a suitable robot grasp given the receiver's upcoming task, and (ii)~a safe yet reachable pose to transfer the object given their arm mobility level. 
    
    \subsubsection*{\textbf{Considering object affordances}} 
        In recent human-to-human user studies, using object affordances has been shown to improve the comfort of the receiver  \cite{cini2019choice,pan2017automated}. These works extract the difference in the giver's choice of grasp when handing over an object arbitrarily in contrast to when considering the receiver's upcoming task. In the latter case, they report a notable preference from the receivers.
        In a robot-to-human setting, there have been fewer explorations of handover methods that directly consider object affordances \cite{chan2014determining,ortenzi2020grasp,aleotti2014affordance}. In \cite{aleotti2014affordance} the authors perform object part segmentation and manually assign the corresponding affordances with the purpose of maximising the user's convenience. \cite{chan2014determining} presents a method where the concept of object affordances for handovers is limited to a one-to-one mapping of object-to-affordance. The authors in \cite{chan2014determining} use human demonstrations and a prior discretisation of grasp configurations to learn a handover model. The resulting model is constrained by the demonstrated affordance. As such, it does not generalise across object classes. \cite{bestick2016implicitly} implicitly uses the concept of affordances for handovers. Authors in \cite{bestick2016implicitly} optimise the receiver's ergonomic cost to grasp the object at a suitable location that facilitates an upcoming task. However, their approach \mbox{does not generalise to unknown objects or tasks.}
    

    \subsubsection*{\textbf{Sharing effort}}
        
        The state-of-the-art in robot-to-human handovers has focused on sharing arm motion effort between the human and the robot \cite{sisbot2012human,pan2017automated,shi2013model,parastegari2017modeling}. One of the key challenges in the field has been to find a location for transfer that balances the receiver's arm comfort \cite{sisbot2012human}, body pose \cite{pan2017automated} and the distance at which the robot's end-effector is considered to be safe \cite{nemlekar2019object,mainprice2012sharing,shi2013model,parastegari2017modeling} from the human body. These works have carried out user studies with participants that do not report arm mobility limitations. Consequently, the preferred location for the transfer results in the human and the robot sharing effort to reach for the object.
        Although these methods are shown to be efficient, they do not offer generalisation guarantees across user populations with potentially different arm movement capacities. 
        
        In contrast to the current literature, we propose an inclusive handover method that adapts to users with high and low arm mobility capacities. To date, no handover technique adapts to low arm mobility levels. Therefore, firstly, we design a method for receivers with low mobility. Secondly, we collect preferences over handover methods from people with different arm mobility capacities. Finally, we learn and generalise such preferences \mbox{across receivers by means of a \ac{SRL}.}

\section{Handover Optimisation with \\ Mobility Constraints \label{sec:costs}}

    We propose a robot-to-human handover method that adapts object configurations to people with different arm mobility. Particularly, we define the handover configurations considering the receiver's (i)~upcoming task, to extract an adequate robot grasp, and (ii)~arm mobility capacities to adapt the object's pose for the transfer. To achieve such a reasoning model (details in Section~\ref{sec:genealising}), first we need to design a method that adapts to people with low arm mobility. Then, we need to analyse preferences over handover methods across users with different arm mobility capacities (Section~\ref{sec:data} presents the online user study). This section details the design of our heuristic-guided hierarchically optimised cost model that adapts handovers to users with low arm mobility.
    
    Current robotic handover methods consider preferences over objects and robot grasp configurations that are not designed for receivers with arm mobility impairments. In contrast, with the insight that less effort means more comfort for the receiver \cite{schweighofer2015effort,sampson2015using}, we model a handover cost that adapts to users with low arm mobility. The heuristic-guided hierarchically optimised cost model extracts (i)~the most suitable robot grasp given the receiver's upcoming task, and (ii)~a transfer object configuration located at a reachable yet safe location for the user. We efficiently guide the configuration search \mbox{through a user-configurable resolution workspace grid map.}
    
\begin{figure}[t!]
          \centering
          \includegraphics[width=8.7cm]{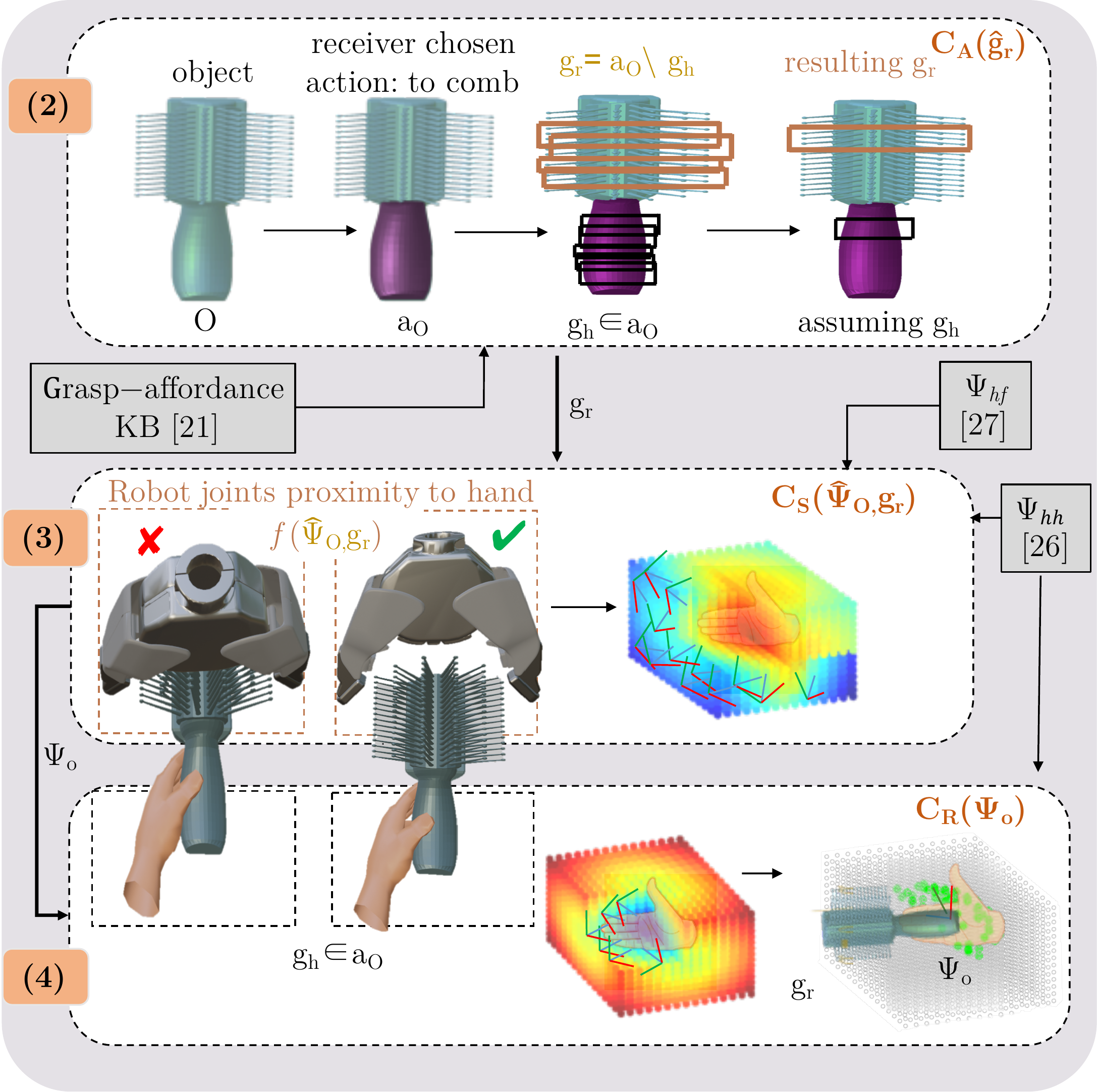}
          \caption{Summary of our heuristic-guided hierarchically optimised cost model. 
          Each block corresponds to (\ref{eq:affordance})-(\ref{eq:reachability}).
          \label{fig:costs}
          }
        \end{figure}

        The resulting map is composed of  $\{\boldsymbol{x},\boldsymbol{y},\boldsymbol{z}\}$ voxels $m_{x,y,z} \in \mathcal{M}_{\{\boldsymbol{x},\boldsymbol{y},\boldsymbol{z}\}}$. Each voxel $m_{x,y,z}$ encapsulates: (i)~non-controllable human values or constants, in our case the human hand $\HHandPose$, face pose $\HFacePose$, and the choice of grasp when receiving the object ${\HumanGrasp}$; and, (ii)~cost-constrained variables which are the configurations we want to optimise, in our case the robot grasp ${\RobotGrasp}$, and object pose $\ObjectPose$. As a result, the map is a function of $ \mathcal{M}_{\{\boldsymbol{x},\boldsymbol{y},\boldsymbol{z}\}}(\RobotGrasp, \ObjectPose)$. We guide the hierarchical optimisation through three costs, as shown in Fig~\ref{fig:costs}. Firstly, we compute an optimal appropriateness cost $\CAffordance$ that gives a suitable robot grasp $\RobotGrasp$ from a set of grasp affordance configurations $\hat{g}_r \in G_r$. Secondly, using the previously found $\RobotGrasp$, we sample for safe object configurations ${\ObjectPose}$ using the safety cost $\CSafety$. This cost is constrained to those object poses $\hat{\Psi}_{O}$ where there is a feasible inverse kinematic solution for the end-effector $\RobotEE$ to proceed with the grasp $\RobotGrasp$, denoted as $f(\hat{\Psi}_{O}, \RobotGrasp) \neq \emptyset$. Finally, in the reachability cost $\CReachability$, we minimise the displacement of the user arm. Given ${\ObjectPose}$, we inform the search for the closest $m_{x,y,z}$ in $\mathbb{R}^3$ and find the optimal object configuration $\ObjectPose$ in SE(3):
        \begin{gather}
          \min_{m \in \mathcal{M}} \CReachability(\ObjectPose)   \nonumber \\
          \resizebox{0.98\columnwidth}{!}{$
          \text{with~} {\ObjectPose} = \argmax\limits_{\hat{\Psi}_{O} \in m} \CSafety(\hat{\Psi}_{O}, \RobotGrasp) 
          \text{~s.t~}  \RobotEE \leftarrow f(\hat{\Psi}_{O}, \RobotGrasp) \neq \emptyset
          $} \nonumber \\
          \text{\small{with}~} \RobotGrasp = \argmax_{\hat{g}_r \in G_r} \CAffordance(\hat{g}_r).
          \label{eq:handover}
        \end{gather}
   
        \textbf{Appropriateness} $\CAffordance(\hat{g}_r)$
            is calculated in the object affordance space and it extracts the grasp configuration the robot should choose depending on the receiver's future task. Depending on the level of human arm mobility impairment, the hand dexterity may vary considerably and, thus, the human choice of grasps. This is a subject worthy of future study. Although we cannot control the human grasp directly, we can leave the object's part that affords the receiver's chosen action occlusion free. Thus, we implicitly offer the receiver the most suitable grasping region. We reason about $\RobotGrasp$, $\HumanGrasp$ and the object affordances regions $a_O$ using the \ac{MLN} \ac{KB} from our earlier work~\cite{ardon2019learning}. The \ac{KB} in \cite{ardon2019learning} is composed of data collected from human users, thus being suitable for the handover task, as well as inferring suitable actions~\cite{ardon2020self}. We consider two sets of grasp configurations: (i)~human grasps are configurations inside the object affordance region $\HumanGrasp \in a_O$, while (ii)~robot grasps are outside, $\RobotGrasp = a_O\setminus \HumanGrasp$. The final goal in (\ref{eq:handover}) is to choose a $\RobotGrasp$ that maximises the distance from the closest possible (i.e., most constraining)~$\HumanGrasp$:
            \begin{equation}
               \resizebox{0.5\columnwidth}{!}{$
                   \CAffordance(\hat{g}_r) =\min\limits_{\HumanGrasp \in a_O}d(\hat{g}_r,\HumanGrasp).
                   \label{eq:affordance}
                   $}
            \end{equation}
            Intuitively, $\CAffordance(\hat{g}_r)$ guides towards appropriate grasps for both, giver and receiver. 

        \textbf{Safety} $\CSafety(\hat{\Psi}_O, \RobotGrasp)$
            is considered in terms of distance between the robot to human. The further away from the human user the robot's manipulator is, the safer it is. Thus, we maximise the distance from the object pose $\hat{\Psi}_O$, projected in $a_O$, to the human hand $\HHandPose$ and face $\HFacePose$, as well as from the $\RobotEE$ to $\HHandPose$. We penalise the cost if any of the distances is below a threshold $t_{h}$ of $5cm$:
            \begin{equation}
                \resizebox{0.99\columnwidth}{!}{$
                    \CSafety(\hat{\Psi}_O, \RobotGrasp) =
                        \begin{cases}
                       d(\hat{\Psi}_O,\HHandPose)+ 
                    d(\hat{\Psi}_O,\HFacePose) + 
                    d(\RobotEE,\HHandPose) ,& \text{if } d(\cdot)\geq t_{h}\\
                        0,              & \text{otherwise.}
                        \end{cases}
                        $}
                        \label{eq:safety}
            \end{equation}
          
        \textbf{Reachability} $\CReachability (\ObjectPose)$
            is introduced to minimise the receiver's arm displacement, thus effort \cite{schweighofer2015effort,sampson2015using}. This cost promotes object configurations located as close to the human hand as possible, consequently, adapting to users with low arm capacities. Specifically, $\CReachability(\ObjectPose)$  penalises the human hand movement from the current pose to the implicitly advised grasp $\HumanGrasp$. \cite{nemlekar2019object} suggested that $75cm$ is a reachable object transfer location, as such, we use it as $t_h$ to penalise greater distances:
            \begin{equation} \resizebox{0.6\columnwidth}{!}{$
             \CReachability(\ObjectPose) =
             \begin{cases}
               d(\HHandPose, \HumanGrasp) ,& \text{if } d(\cdot)\leq t_h\\
                    \infty,              & \text{otherwise.}
                \end{cases}
                $}
                \label{eq:reachability}
            \end{equation}
    
        In summary, using (\ref{eq:handover}), the robot obtains the most appropriate robot grasp given the receiver's task and a safe yet reachable object configuration. As a result, adapting handovers to users with low mobility impairments. Fig.~\ref{fig:costs} summarises the heuristic-guided hierarchically cost model.

\section{User Handover Preferences}\label{sec:data}

    To implement an inclusive handover method that adapts to people with different arm mobility levels (Section~\ref{sec:genealising}), we need to explore users preferences on handover methods. To identify such preferences, we feature different handover methods, including ours (as detailed in Section~\ref{sec:costs}), and present them to users through an online study. In this section, we describe our user study setup, hypothesis on preferences, data collection and a systematic evaluation of the users' perception.

    \subsection{User Study Setup \label{sec:setup}}

\begin{figure}[t!]
            \centering
            \adjustbox{trim=1cm 0 2cm 0}{%
                    \includegraphics[width=8.5cm]{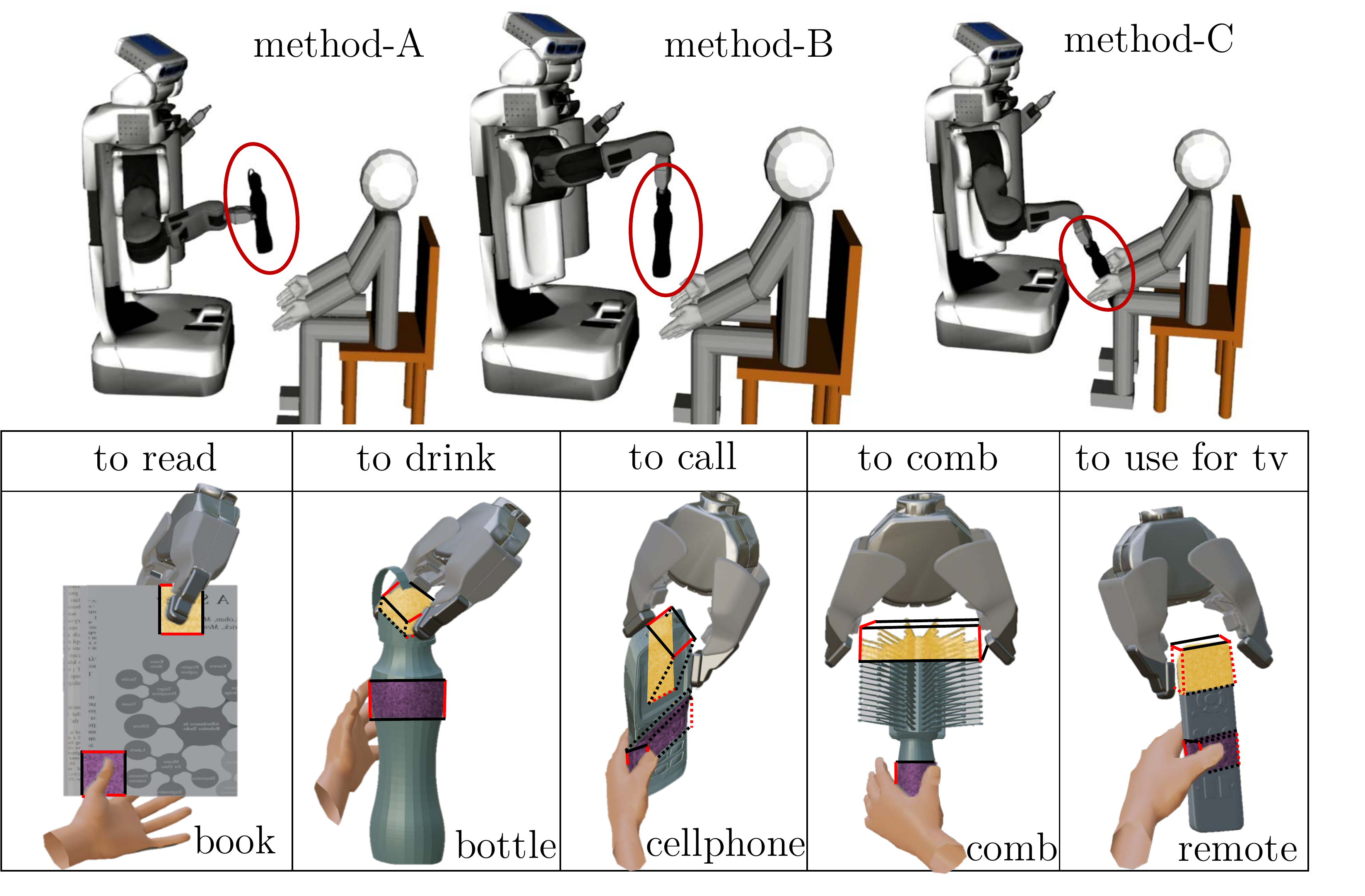}
                }
            \caption{First row: video frame samples of the methods presented to the users. Second row: objects in the online study with $\HumanGrasp$ (purple) and $\RobotGrasp$ (ochre) as detected with our method. \label{fig:online_design}}
        \end{figure}

        For the user study we consider three different handover methods. As \textit{method-A}, we implement a handover technique following the guidelines in \cite{nemlekar2019object,mainprice2012sharing}. These works extract the optimal object transfer point. As in \cite{nemlekar2019object,mainprice2012sharing}, for method-A we set the object transfer point at a distance of $75cm$ from the human body and an arbitrary robot grasp. As \textit{method-B}, we use \cite{parastegari2017modeling}'s suggested transfer location at $50cm$ and a robot grasp that considers the receiver's upcoming task. As \textit{method-C}, we use our proposal in Section~\ref{sec:costs}. As a result, the three different methods handover objects in three different ways. The first row of Fig.~\ref{fig:online_design} shows an example of an object's final pose for each handover method. To the participants, neither the methods' details nor name were disclosed. For the reminder of this manuscript \textit{method-C} will be referred to as \textit{ours}.

    \subsection{Premise on Handover Preferences}

        We hypothesize that the preferences of robot-to-human handovers vary according to the human's level of arm mobility. We argue that people's preferences could be suitable described by one of the previously presented handover methods. We base this premise on the theoretical grounds that the \textit{effort} required to move the arm joints will be optimized. \cite{schweighofer2015effort} uses mathematical models to calculate joints effort alongside average measures of the human body. We use \cite{schweighofer2015effort}'s method to design and simulate the kinematics and dynamics of a human. To calculate the effort, we consider variations of \ac{DoF} of the main arm joints: shoulder, elbow and wrist. We run $5$ trials per handover method (i.e., method-A, method-B and ours), and calculate the effort on the simulated human joints. Considering $3$ different handover setups, the average effort results in (i)~$39~Nm$ for method-A, (ii)~$20~Nm$ for method-B and (iii)~$12~Nm$ for ours. The values show our proposal eases arm effort. Furthermore, there are less tangible, or qualitative, but equally important benefits of usability in the proposed approach. However, they are not enough to assume preferences for the handover setups. To study user preferences we run an online survey.

    \subsection{Data Collection \label{sec:design}}

        We collected our data through an online survey to guarantee social distancing rules to our participants\footnote{Heriot-Watt University ethical approval: MACSPROJECTS 2184}.
        Contrary to previous works, our goal is to achieve an inclusive robot-human handover technique. This was done in collaboration with \ac{CHSS}\footnote{Health charity supporting people across Scotland with rehabilitation given chest, heart and stroke conditions. \url{https://www.chss.org.uk/}} to recruit participants that suffer from arm mobility impairments. Through \ac{CHSS}, we recruited a total of $9$ volunteers. Additionally, we used Amazon’s Mechanical Turk platform, to obtain opinions from people with varied arm mobility capacities. From Mechanical Turk, we obtained a total of $250$ unique participants, using the following criteria: English proficiency, approval rate, filtering questions, and task time. Participants were paid $\$2.50$, an amount on par with the going rate at the time for online surveys of $\approx$20 min duration.

        We hypothesise that people with different arm mobility capacities prefer different types of robot-human handovers. Our goal with the online study is to learn people's preferences on different handover setups including our proposal in Section~\ref{sec:costs}.
        The participants were given a consent form and a three pages questionnaire. \textit{Firstly,} we requested demographics, including an animation to identify their arm mobility capacities following the suggestion from \cite{schweighofer2015effort}. The resulting $259$ participants ($171$ males and $87$ females) were aged between $18$ and $69$ years old with only $10$\% ($26$ participants) of the population being familiar with robots. $27.4\%$ of the sampled population ($71$ participants) reported some level of arm mobility impairment and associated it with one of the animations presented in the survey.

        \textit{Secondly,} we presented to each participant $3$ different short clips. Each one of the three short clips showed a different handover method, as explained in Section~\ref{sec:setup}. The $3$ clips were randomised among $5$ objects. The $5$ objects are of common use on \ac{ADL} by people with \ac{ALS} \cite{choi2009list}. The second row of Fig.~\ref{fig:online_design} shows the objects, their upcoming task and corresponding robot grasp used in the online study. In this illustration, $\HumanGrasp$ (purple patch) is our hypothesis of the most likely choice of human grasp. Each clip was approximately $40~s$ in length and followed by a 5-point scale that measured: safety, comfort and appropriateness of the handover technique. The order of the three clips varied randomly across participants to avoid biasing the 5-point scale metric. \textit{Finally,} we asked the participants (i)~to rank the importance of factors such as the effort to reach the object, naturalness and appropriateness of robot grasp, (ii)~to select preferred technique overall, and (iii)~an open-ended question about their opinion on robot-human handovers. In the end, we debriefed them on the purpose of the study. In summary, we acquired data from a total of $259$ participants distributed among $15$ different setups. Namely, $5$ objects with $3$ methods for each handover setup, $2$ representing state-of-the-art and ours.

        \subsection{Systematic Analysis of User Input \label{sec:user_responses}}

            \begin{figure*}
                \centering

                \includegraphics[width=17.8cm]{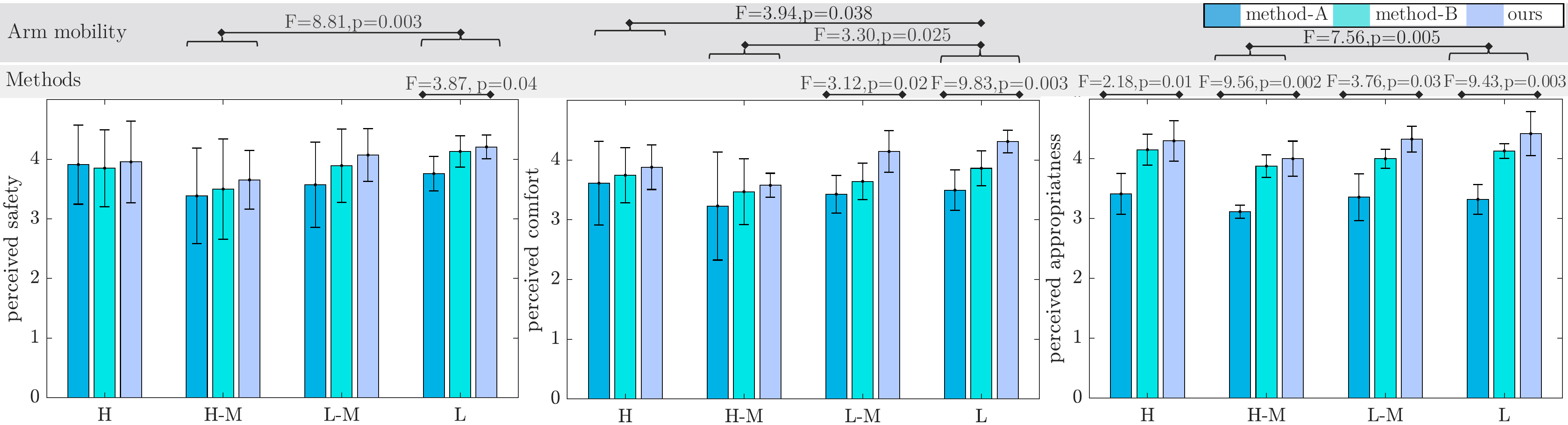}

                \caption{Mean and variance of participants perception on safety, comfort and appropriateness in a 5-point scale metric. Evaluation includes two-way repeated ANOVA. Only significant statistical results are shown, i.e., with a $p< 0.05$.
                \label{fig:data_collection}}
                \vspace{-0.3cm}
            \end{figure*}


        We examine the participants' responses to detect handover preferences. Guided by our hypothesis, we analyse the data to show the influence of arm mobility level and handover technique interaction. Fig.~\ref{fig:data_collection} shows a summary of the findings as extracted from the 5-point scale metric. The higher on the scale the safest, more comfortable or appropriate the handover method is. As mentioned in Section~\ref{sec:design}, we create animations for the users to identify their arm mobility level. The participants identified themselves in either of the $4$ shown animations: high (H), high-medium (H-M), low-medium (L-M), and low (L) arm mobility. For each of the levels, we illustrate the mean and standard deviation of the three handover methods included in the study.
        Our analysis involves a normally distributed two-way repeated-measures \ac{ANOVA}, using the handover methods and users arm mobility levels as factors, and participant ID as repetitions.

        \begin{table}[b!]
            \renewcommand{\arraystretch}{1.25}
             \centering
             \begin{tabular}{|c|c|c|c|c|}
             \cline{2-5}

              \multicolumn{1}{c|}{} & \multirow{2}{1.5cm}{\textbf{participants}} &  \multicolumn{3}{c|}{\textbf{Handover method preference overall}}  \\ \cline{3-5}
              \multicolumn{1}{c|}{} & & \textbf{method-A} & \textbf{method-B} & \textbf{ours} \\ \hline

              \!\!\!H & 179 & 23.5\% & \textbf{73.7\%} & 2.8\% \\ \hline
              \!\!\!H-M & 27 & 23.1\% & \textbf{65.4\%} & 11.5\% \\ \hline
              \!\!\!L-M & 18 & 7.1\% & 28.6\% & \textbf{64.3\%} \\ \hline
              \!\!\!L & 35 & 3.2\% & 16.1\% & \textbf{80.7\%} \\ \hline

             \end{tabular}
             \caption{Distribution of participants per arm mobility level as they chose their preferred handover method.
             \label{tb:preference}}
        \end{table}

        For \textit{perceived safety}, there is no significant difference across methods as rated by users in groups H and \mbox{H-M}. Receivers in \mbox{L-M} scored our method as slightly safer. Nonetheless, only users in \mbox{L} scaled method-A as significantly different from ours. The arm mobility of the users influences different safety perceptions on users from \mbox{H-M} to those in L. In \textit{perceived comfort}, all users perceived our method as the most comfortable one. Namely, for \mbox{L-M} and L, comfort is significantly different between method-A and ours. In terms of arm mobility, levels H and \mbox{H-M} have significantly different perceptions of comfort from the users who identified themselves at L level. For \textit{perceived appropriateness} of the robot grasp given the user upcoming task, there is a significant difference across method-A and ours for all arm levels. The gap between the method-A and the other two methods is noticeable. On average, our method is perceived positively by the users. Nonetheless, the preference for our method over the other two setups is clearer in participants that reported lower levels of arm mobility (i.e., \mbox{L-M} and L). Users belonging to \mbox{H-M} have a significantly different perception of appropriateness from those users at the L level.
        Moreover, Fig.~\ref{fig:data_collection} results go in accordance with the findings of \cite{cini2019choice,pan2017automated} where participants, in general, prefer the techniques that consider the receiver's upcoming task.
        In this regard, we asked the participants to choose their overall preferred handover method. Table~\ref{tb:preference} shows a summary of the preference distribution as related to arm mobility levels.
        \begin{figure}[b!]
            \centering
            \adjustbox{trim=0cm 0 0cm 0}{%
                    \includegraphics[width=7.5cm]{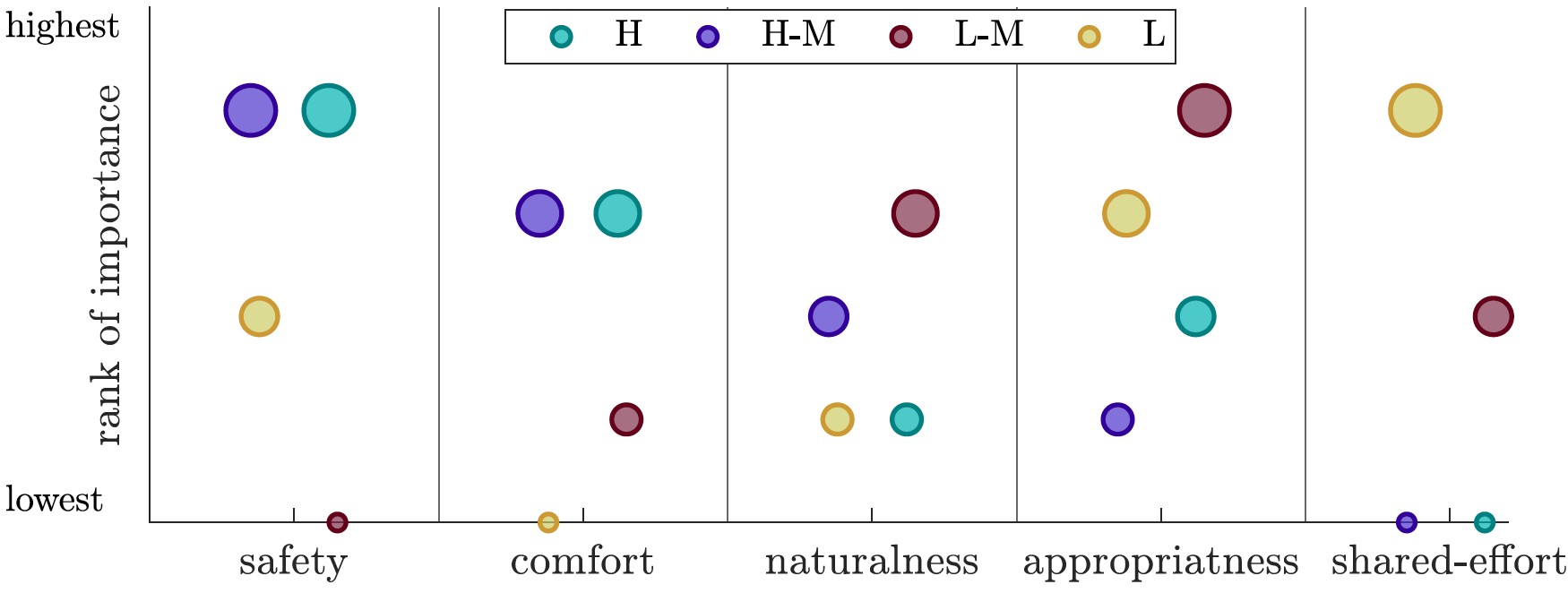}
                }
            \caption{Median rank consensus per arm mobility of the considered aspects that influence a handover task.
            \label{fig:importance}}
        \end{figure}

        The participants also ranked, from most to least important, the following aspects: (a)~safety, (b)~comfort, (c)~naturalness of the handover, (d)~appropriateness of the robot grasp given the receiver's upcoming task, and (e)~that the robot moves more than the human to reach the object transfer location (i.e., shared effort). Fig.~\ref{fig:importance} illustrates the median rank consensus per arm mobility level. The resulting ranking demonstrates the difference in priorities, especially on the extremes of the arm mobility level spectrum. For example, for H and \mbox{H-M} feeling safe and comfortable is the top priority. In contrast, for \mbox{L-M} and L the preference fluctuates between the robot moving more than the human to transfer the object and obtaining the object in a configuration that they can easily use afterwards. As in Table~\ref{tb:preference}, Fig.~\ref{fig:importance} reiterates that users with lower arm mobility prefer a technique that brings the object closer to the hand.

          \begin{table*}[t!]
            \renewcommand{\arraystretch}{1.25}
             \centering
             \begin{tabular}{|c|p{0.35cm}|p{7.2cm}||p{0.35cm}|p{7.2cm}|}
             \cline{2-5}

              \multicolumn{1}{c|}{} & \textbf{\%} & \textbf{ \centering Sample of positive responses} & \textbf{\%} & \textbf{Sample of negative responses} \\ \hline

              \!\!\!H & 23.5 &  ``... \textsc{as long as} it is safe \textsc{I would be comfortable''} & 24 & ``It \textsc{depends} ... because of the risk of something happening that the robot cannot \textsc{adapt}''\\ \hline

              \!\!\!H-M & 57.7 & ``\textsc{ I'd feel comfortable if} the robot's help is convenient ...'' & 7.7 & ``It \textsc{depends} on the object and the setting''  \\ \hline

              \!\!\!L-M & 50 & ``\textsc{... I feel comfortable,} it behaves \textsc{usefully}'' & 7.14 & ``... make sure that I am able to override its functions with voice commands if it malfunctions or behaves \textsc{unexpectedly}'' \\  \hline

              \!\!\!L & 64.5& ``It looks \textsc{comfortable}. I am the primary care taker of my sister. This could be really \textsc{useful} for her ...'' & 3.32 & ``I \textsc{would only} be worried about dangerous objects'' \\  \hline

             \end{tabular}
             \caption{We detect sets of words with higher recurrence that hint positive and negative responses per arm level. The $\%$ indicates the appearance events of the \textsc{keywords} sets.
             \label{tb:open_ended}}

              \vspace{-0.3cm}
         \end{table*}

        Finally, we asked in an \textit{open-ended question} how the users feel about robot-human handover collaboration tasks. Table~\ref{tb:open_ended} shows a sample set of responses. Per arm level, we created a word count of the responses and extracted sets of words appearing with higher frequency. Some of the extracted sets imply a positive opinion about the task, while others suggest a negative or doubtful perception of the robot's performance. For example, in the set implying positive perceptions, the keyword \textsc{comfortable} is used by participants in all levels. Nonetheless, for H  and \mbox{H-M} \textsc{comfortable} is often found along \textsc{safe} and a conditional such as \textsc{if} or \textsc{as long as}. On the other hand, for \mbox{L-M} and L the appearance of \textsc{comfortable} is followed by \textsc{useful} or \textsc{usefully}. By putting these words in context, it is clear that, depending on the mobility level, some participants accept the collaboration with reservations while others perceive the robot as a helper. The participant's statements support the ranking on Fig.~\ref{fig:importance}.

        In summary, although users in general prefer a method that considers their upcoming task, there are different preferences related to user arm mobility capacities. Receivers with low levels of arm mobility prefer the robot to perform most of the handover task, while users with high mobility choose to have some freedom and share the task effort.

\section{Generalising Handovers to New Objects}\label{sec:genealising}

 We aim to generalise handover configurations to previously unseen objects, subject to the users' upcoming task and arm mobility capacities. We use preferences collected in Section~\ref{sec:data} to expand a previously proposed relational model in \cite{ardon2019learning}. This section presents the design, execution and evaluation of the relational policy which encapsulates users' preferences over a variety of handover techniques. The resulting policy generalises handover preferences onto semantically similar objects across users with different arm mobility.

    \subsection{Statistical Relational Learner \label{sec:learner}}

        As explained in Section~\ref{sec:data}, we use different handover methods, including ours in Section~\ref{sec:costs} as prior to collect user preferences. Our goal with the collected data is to create a handover reasoning model for different objects considering the receiver's arm mobility level and upcoming tasks. To achieve this, we need a methodology that ensembles different structural features such as object affordances, grasp configurations and preferred object configurations given a level of arm mobility. To this end, we apply an \ac{SRL}, in particular \ac{MLN} \cite{richardson2006markov}. Previously in \cite{ardon2019learning}, we used a pre-trained \acp{CNN} to extract object semantic features and linked them to grasp affordances using \ac{MLN} to build a \ac{KB}. To include handover preferences in the \ac{KB}, we need to relate the existing semantic descriptors to new structures representing object configurations and user arm mobility capacities.

        Formally, a \ac{MLN} is a set of formulae $f \in \boldsymbol{F}$ in \ac{FOL} and real-valued weights $w_i$ attached to each~$f$. The probability distribution over the set of ground truth associations, i.e., collected data $X$ is:
        \begin{equation}
            P(X=x) = \frac{1}{Z} \exp \left({\sum_i w_i n_i(x)} \right)\!,
        \end{equation}

        where $x$ is an instance of the data, in our case each participant's set of responses; $n_i(x)$ is the number of observations supporting a formula of $f$ in $x$, and $Z$ is a normalisation constant. We learn the optimal weights $w^*$ from maximising the pseudo-log-likelihood \mbox{$log P^*_w(X=x)$} of the obtained probability distribution of the available $X$. In Section~\ref{sec:costs}, we use the grasp affordance policy in \cite{ardon2019learning} as a prior to obtain suitable robot grasps considering the receiver's upcoming task. Using the (i)~resulting grasps from $\CAffordance(\hat{g}_r)$, (ii)~the object affordance semantic descriptions collected in \cite{ardon2019learning}\footnote{Found in: \url{http://bit.ly/semantic_features}} (i.e., object categories, visual description of shape context, texture and material), and (iii)~the preferences from Section~\ref{sec:user_responses}, we create instances $x$ that represent the connecting entities we want to apply probabilistic and logical reasoning to. Intuitively, the object configurations resulting from the variations of the heuristic-guided model can be regarded as entries-to-labels in a relational database.
        From \cite{ardon2019learning}, we deduced that clustering the objects by shape context semantic description offered generalisation opportunities to create grasping hypotheses. Using this observation, we average the object poses $\ObjectPose$ and their corresponding object's robot grasp $\RobotGrasp$ per shape context. As a result, the \ac{SRL} generalises object configurations to previously unseen but semantically similar objects.

\begin{figure}[b!]
          \centering
          \includegraphics[width=8.6cm]{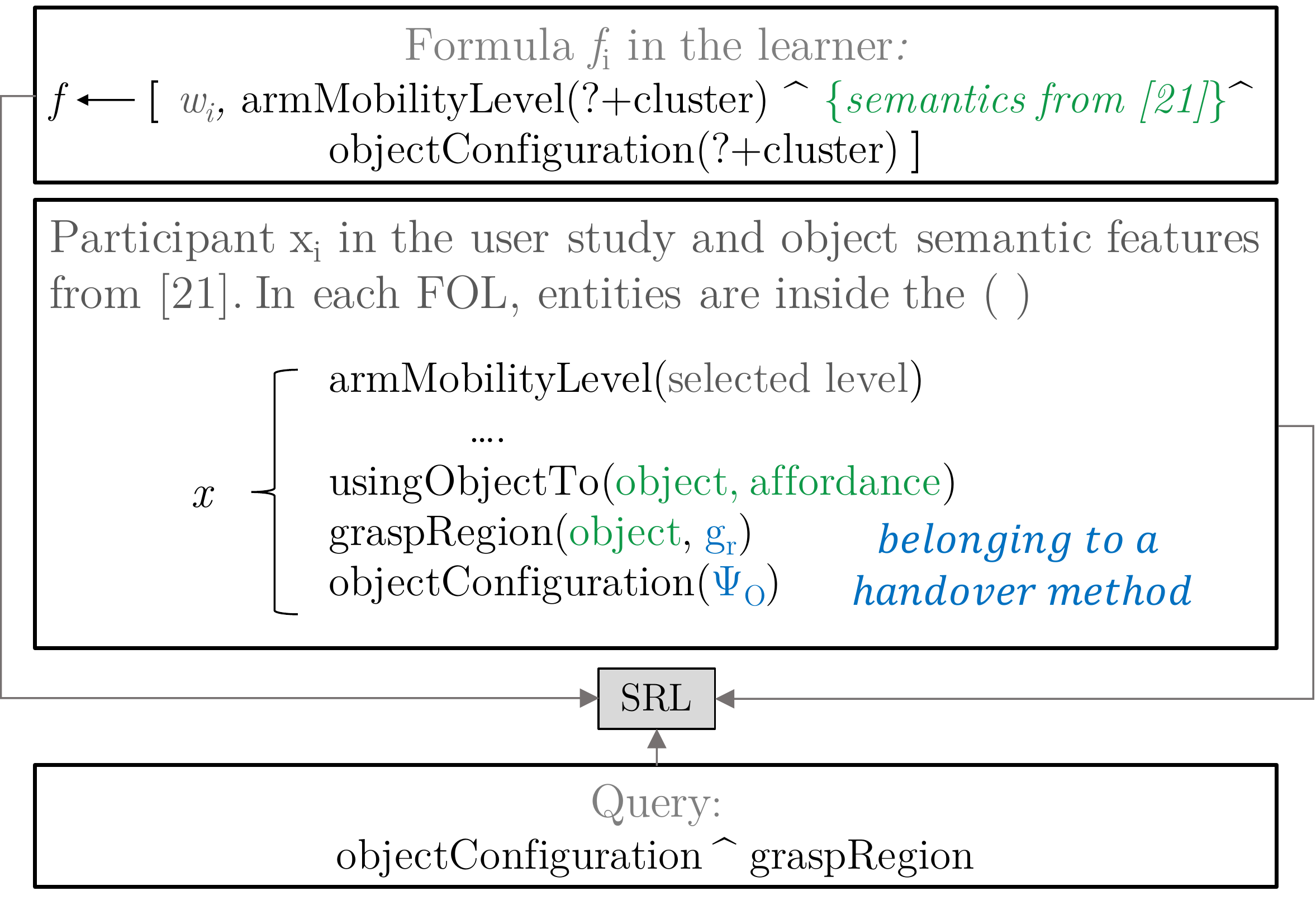}
          \caption{Summary of the \ac{SRL} and its components.\label{fig:srl}
          }
        \end{figure}

        Fig.~\ref{fig:srl} exemplifies the \ac{SRL}, where the preferred handover object configuration (objectConfiguration) and the robot grasp on the object (graspRegion) are the entities we want to predict. We use \ac{WCSP} \cite{jain2009markov} to infer these two parameters given the set of entities in a query. Using the resulting optimal object and grasp configuration, the robot is able to generalise the handover method to different objects depending on the query entities.

    \subsection{Execution \label{sec:rollout}}

        Algorithm~\ref{alg:framework_handover} presents an outline of the handover end-to-end execution. The algorithm aims to provide a robotic platform with a feasible handover object configuration given the receiver's arm mobility level and upcoming task.
        From the user, we obtain the desired upcoming task to perform with an object in the scene and the arm mobility level the receiver identifies to (line~\ref{alg_line:input1} to \ref{alg_line:input2}). The \ac{SRL} trained model in Section~\ref{sec:learner} is fed with the user's information (line~\ref{alg_line:input3}). Given the visual perception and extraction of the objects semantics \cite{ardon2019learning}, human hand pose \cite{chen2019pose} and human face pose \cite{breitenstein2008real} (line~\ref{alg_line:input4} to \ref{alg_line:human_pose}), the end-to-end execution is as follows. The model infers the optimal object transfer configuration OTC$^*$ composed of the appropriate robot grasp for the user's upcoming task, and the suitable object pose given the user's arm capacities (line~\ref{alg_line:kb}). Second, on the suitable robot grasp we calculate a safe grasping configuration $\RobotEE$  (line~\ref{alg_line:ee_pose}). Finally, we allow the robot to move to the object transfer pose as long as it keeps the safety threshold from the human, as in (\ref{eq:safety}), (\ref{eq:reachability}) (line~\ref{alg_line:while} to \ref{alg_line:send_transfer_goal}). All \mbox{the configurations are in the robot's workspace.}

        \begin{algorithm}[t!]
            \caption{end-to-end execution \label{alg:framework_handover}}
            \textbf{Input:} \\

           \textit{affordance}: user's defined upcoming task \\ \label{alg_line:input1}
            \textit{level}: user's identified arm mobility level \\ \label{alg_line:input2}
            $\texttt{Handover}$: SRL handover model (\textit{affordance}, \textit{level}) \\
            \label{alg_line:input3}
            $\text{\textit{CP}}$: camera perception \\ \label{alg_line:input4}
            $\texttt{extractSemantics}$: DCNN from \cite{ardon2019learning}  \\
            $\texttt{trackHumanPose}$: CNNs from \cite{chen2019pose} and \cite{breitenstein2008real} \\

            \Begin{
                $\textit{semantics} \gets$ \texttt{extractSemantics}(\textit{CP.in2D})\\ \label{alg_line:semantics}
                $\textit{h} \gets$ \texttt{trackHumanPose}(\textit{CP.in3D})\\ \label{alg_line:human_pose}
                \textit{OTC$^*$} $\!\gets\!$ \texttt{Handover}(\textit{affordance, level, semantics})\\ \label{alg_line:kb}
                \textit{$\RobotEE$} $\gets$ \texttt{safeRobotGrasp}(\textit{OTC})\\ \label{alg_line:ee_pose}
               \While{ \textup{\texttt{safeDistanceFrom}(\textit{h.hand, h.face})}} {\label{alg_line:while}
               \If{\textup{\texttt{objectGrasped}(\textit{$\RobotEE$})}}{ \label{alg_line:send_ee_pose}
                    \texttt{robotMoveTo}(\textit{OTC$^*$}) \label{alg_line:send_transfer_goal}
                    }
                }
            }
            \vspace{-0.1cm}
        \end{algorithm}

    \subsection{Reasoning about Handover Tasks}

        In the last stage, we evaluate the generalisation of the learned handover reasoning policy\footnote{\mbox{Code available: \url{http://bit.ly/handover_sim}}}. To assess the learned model, we start by analysing the differences in robot grasps resulting from reasoning about grasp affordances with \cite{ardon2019learning} versus our proposal in Section~\ref{sec:costs} for handover. Moreover, we evaluate the generalisation performance of our handover reasoning model with semantically similar objects, given the user arm mobility level and upcoming task.

        \subsubsection*{\textbf{Grasp affordance reasoning}}
            We are interested in evaluating the dissimilarity in the choice of robot grasp when the robot (i)~is detecting the object grasp affordance to solely manipulate the object, as in \cite{ardon2019learning}, in contrast to (ii)~when it is identifying the object grasp affordance to accommodate the receiver's upcoming task, i.e., our proposal in Section~\ref{sec:costs}. For this evaluation, we consider the 2D image of $5$ object instances, with one affordance, for each of the $32$ objects in \cite{ardon2019learning}'s \ac{KB}. For simplicity purposes, as in \cite{ardon2019learning}, we group the objects by shape context in the following illustrations.

\begin{figure}[t!]
                    \centering
                    \adjustbox{trim=0cm 0 0cm 0}{%
                        \subfloat[Rank sum test. Threshold set at $p=0.05$\label{fig:w_test}]{
                           \includegraphics[width=\columnwidth]{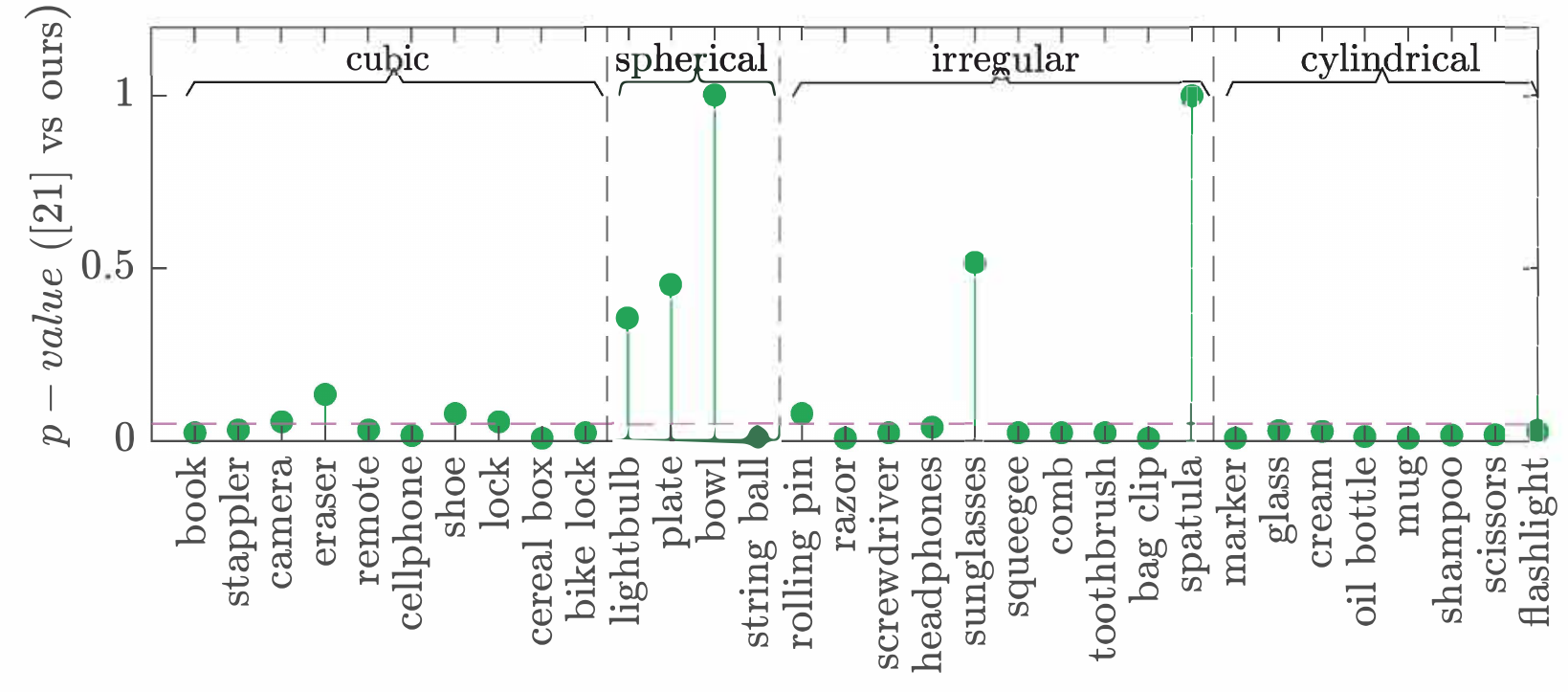}
                        }
                    } \vspace{-0.5cm}

                    \begin{minipage}{0.40\textwidth}
                        \begin{minipage}{0.25\textwidth}
                          \adjustbox{trim=0.78cm 0cm 0cm 0cm}{
                            {
                                \includegraphics[width=6cm]{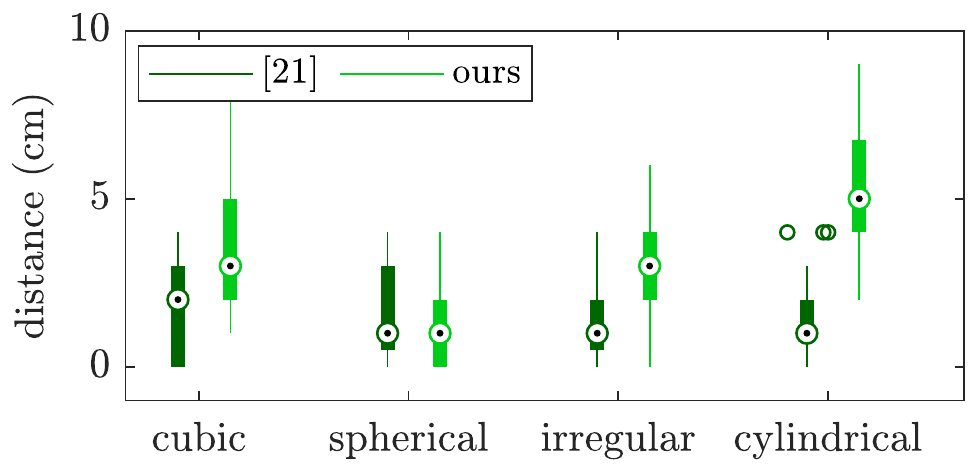}
                            }
                        }
                        \end{minipage}
                        \begin{minipage}{0.15\textwidth}
                            \adjustbox{trim=-3cm -1.5cm 0cm 0cm}{
                             \includegraphics[width=2.8cm]{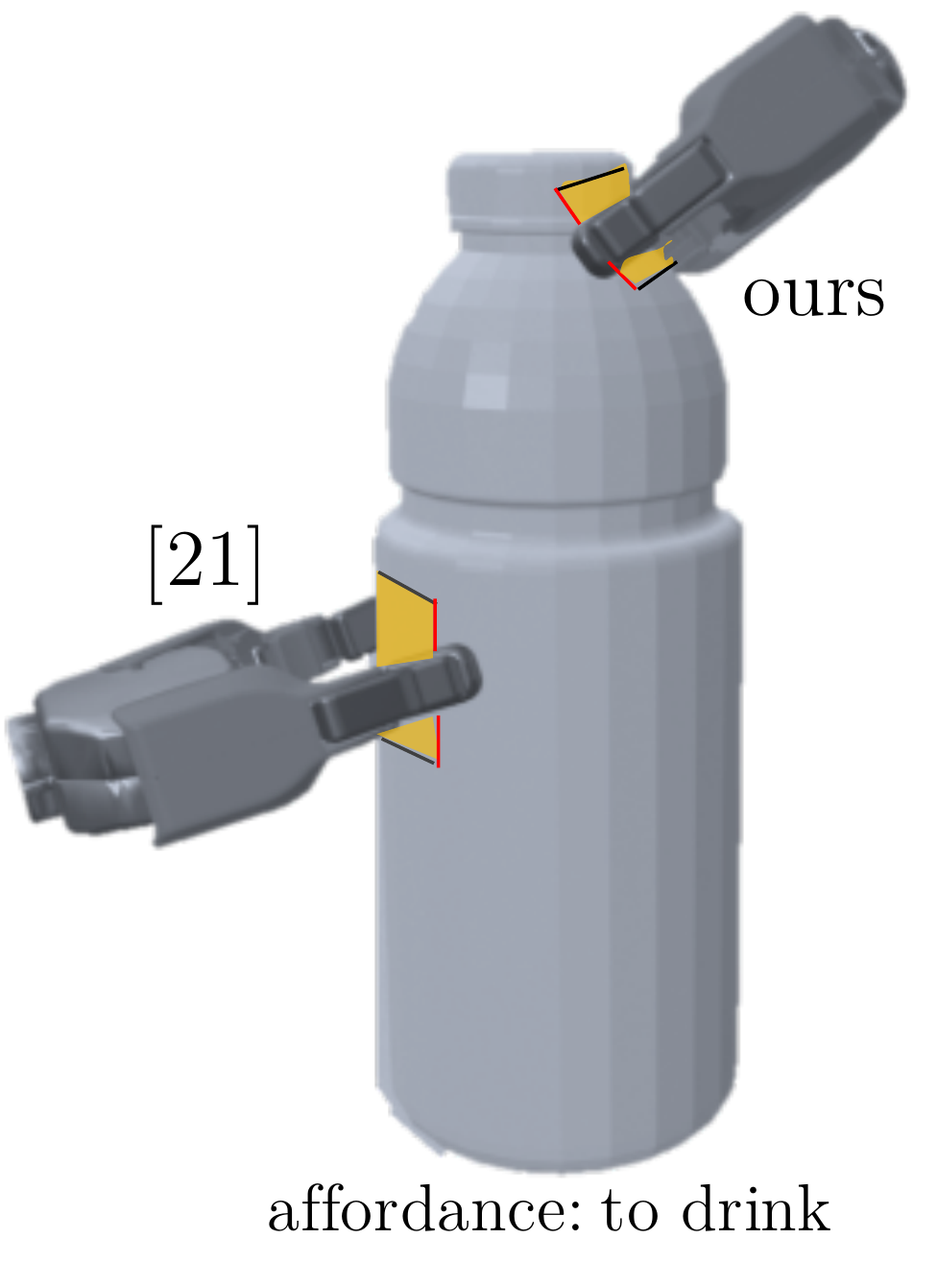}

                            }
                        \end{minipage}

                    \vspace{-1.6cm}
                    \subfloat[Grasp distribution as distance from the object centre \label{fig:grasp_distribution}]
                    {
                        \includegraphics[width=\columnwidth]{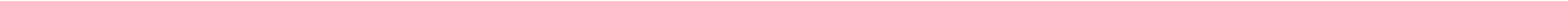}
                    }
                    \end{minipage}
                  \caption{Difference on robot grasp distributions when calculating the grasp region with \cite{ardon2019learning} vs. ours.
              \label{fig:p_sginificance}}
                        \label{fig:p_sginificance_grasps_distribution}
            \end{figure}

            Firstly, we calculate the rank-sum test for the different robot grasp choices between (i) and (ii) to evaluate if those choices are significantly different. Fig.~\ref{fig:w_test} shows a summary of the calculated p-values. Second, we calculate the average Euclidean distance from the centre of the objects to the obtained robot grasp. Fig.~\ref{fig:grasp_distribution} shows the calculated distance in $cm$. In Fig.~\ref{fig:p_sginificance_grasps_distribution} all the objects, except homogeneous shaped ones, show variations on the robot grasp. Interestingly, for handing over, the choice of grasp is located towards the objects extremes. Therefore, showing that the system reasons on different robot grasp when tasked to handover objects.

        \subsubsection*{\textbf{Handover reasoning}}
\begin{figure}[t!]
            \centering
            \adjustbox{trim=0cm 0cm 0cm 0}{%
                \subfloat[remote control, mobility: H\label{fig:to_drink}]{
                   \includegraphics[width=4.0cm]{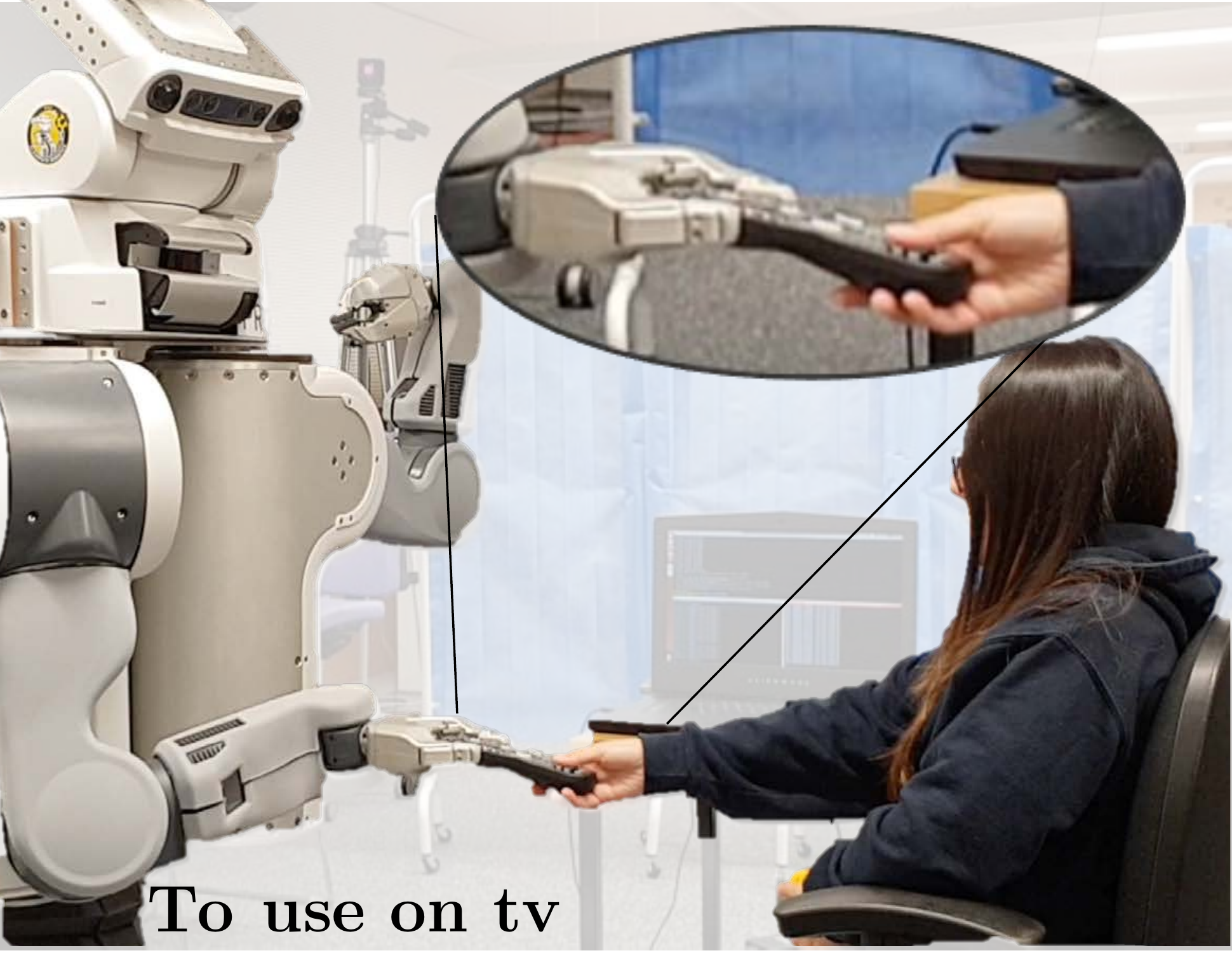}
                }
            }
             \adjustbox{trim=0.0cm 0cm 0cm 0cm}{%
                \subfloat[hair comb, mobility: L\label{fig:to_place_aside} ]{
                    \includegraphics[width=4.0cm]{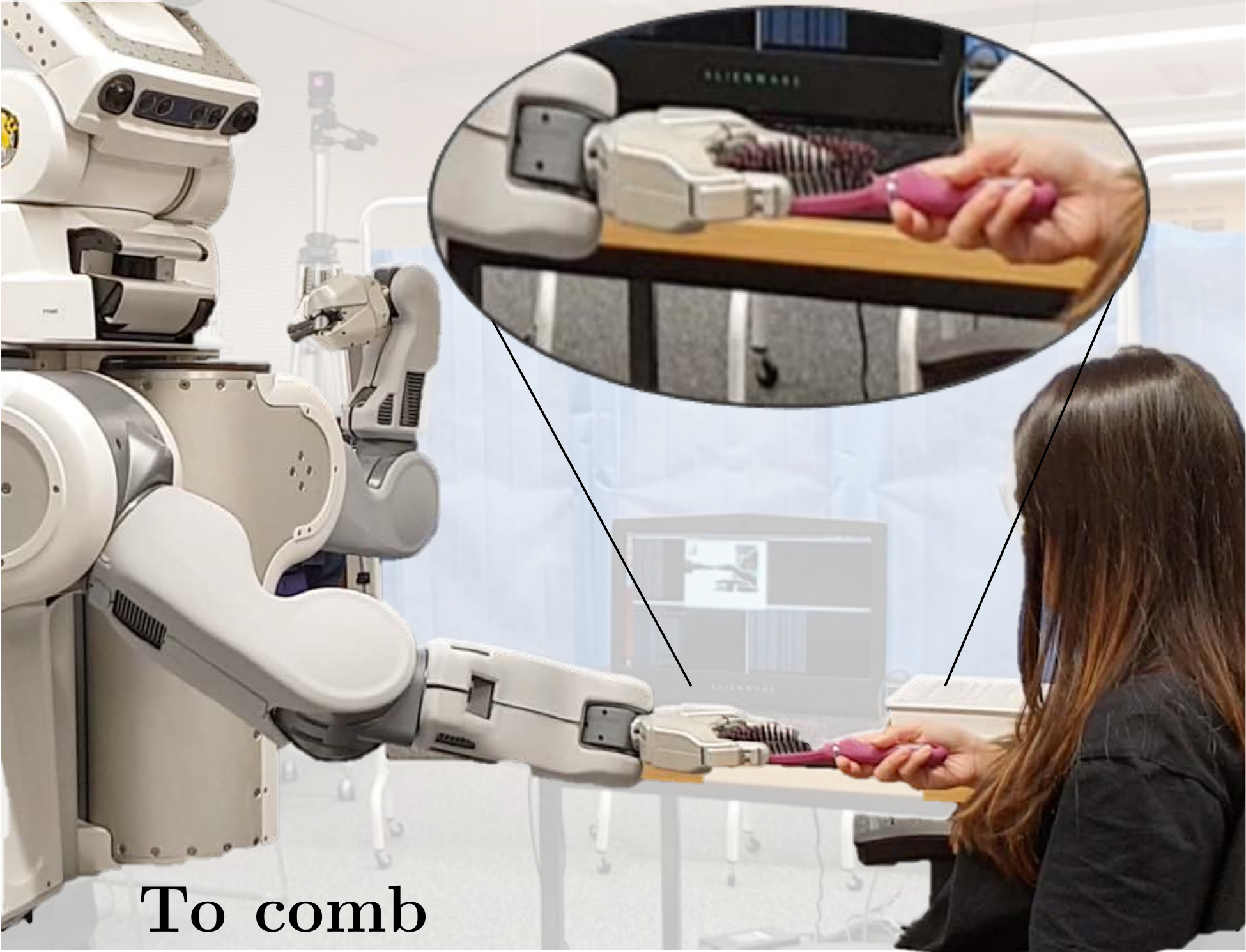}
                }
            } \vspace{-0.25cm}

            \adjustbox{trim=0.2cm 0 0cm 0}{%
                \subfloat[bowl, mobility: H-M\label{fig:to_drink}]{
                   \includegraphics[width=4.0cm]{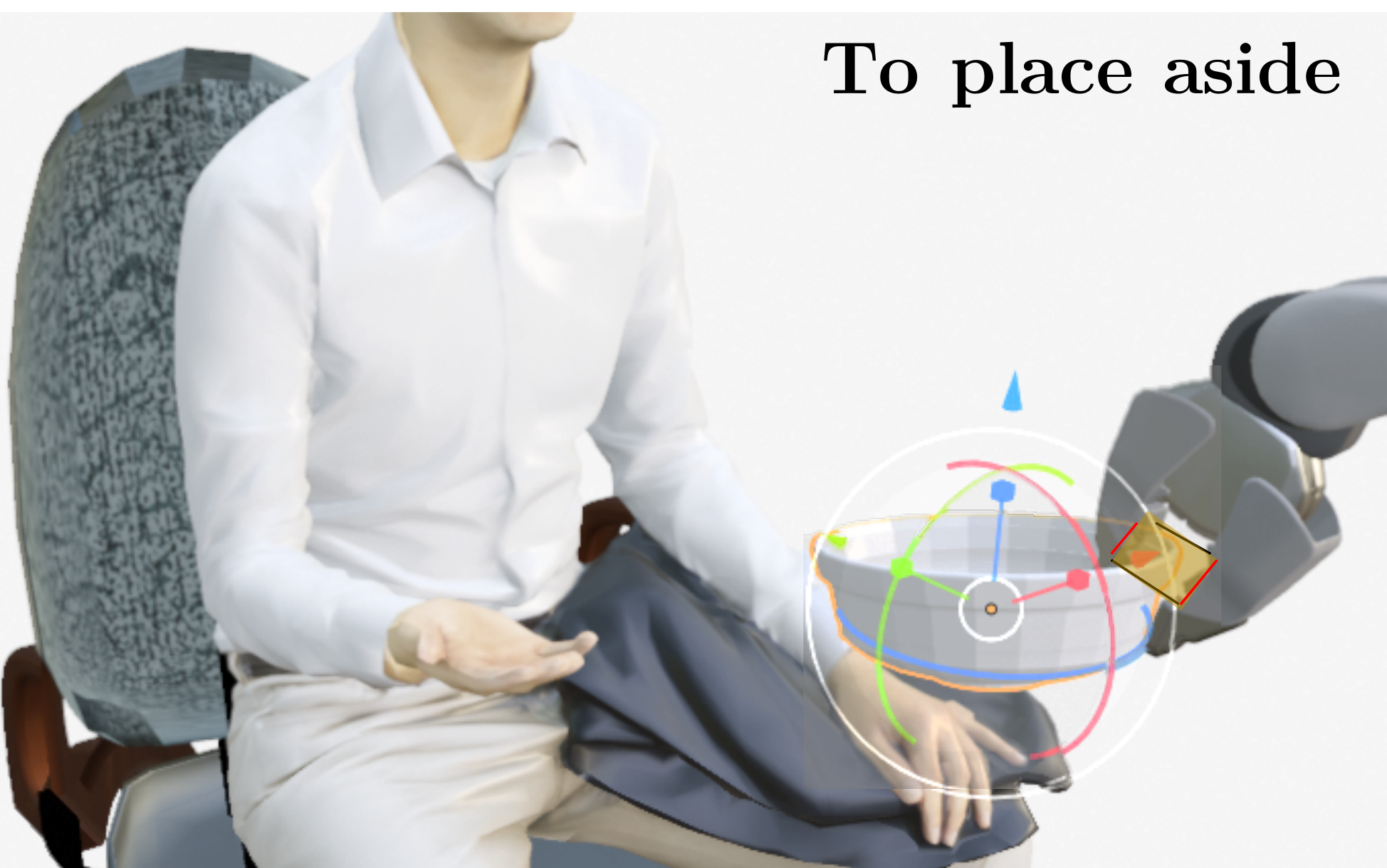}
                }
            }
             \adjustbox{trim=0.0cm 0cm 0cm 0cm}{%
                \subfloat[glass, mobility: L-M\label{fig:to_place_aside} ]{
                    \includegraphics[width=4.0cm]{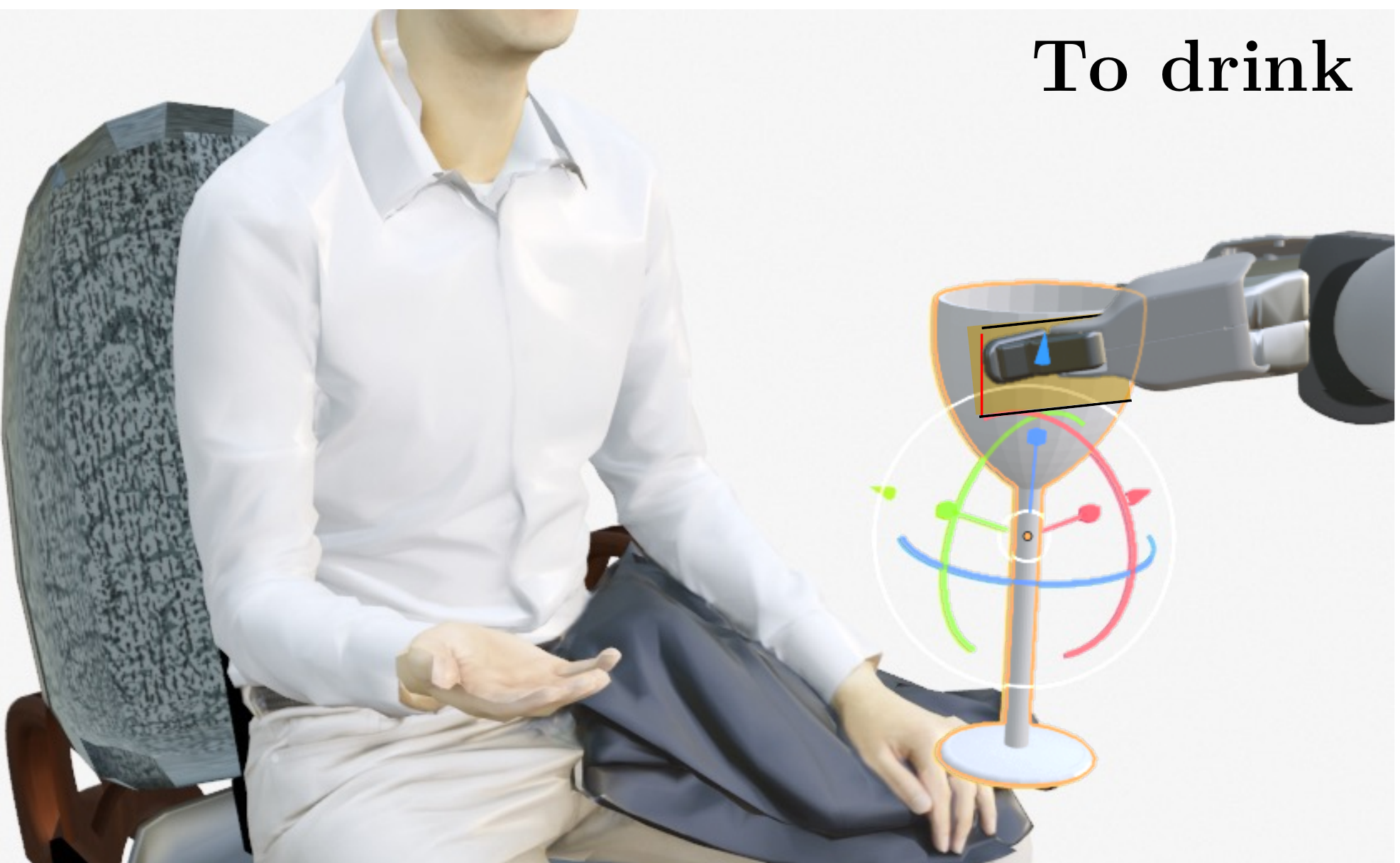}
                }
            }
            \vspace{-0.1cm}

          \caption{Examples of the PR2 handing over known (a)-(b) and unknown objects (c)-(d), in real and simulated scenarios, using our \ac{SRL} model.
      \label{fig:experiments}}
      \vspace{-0.25cm}
    \end{figure}

            We test if the \ac{SRL} generalises handovers to semantically similar objects subject to arm mobility levels and the receiver's upcoming task. For this purpose, we extend the $259$ ground truth data instances of the $5$ objects obtained from the online study. Using the user preference distribution of Table~\ref{tb:preference} on the object shape context (as in Fig.~\ref{fig:w_test}), we synthetically generate $1{,}398$ handover configurations of $27$ other objects. In total, the extended dataset contains $1{,}657$ instances across $32$ objects. We feed to the \ac{SRL} $70\%$ ($22$) of the objects for learning and leave $30\%$ ($10$) semantically similar objects for testing. As shown in Table~\ref{tb:grasp}, we obtain an overall average handover pose ($\pm 5mm$, $\pm 2^\circ$) accuracy of $90.8\%$ on the testing set when inferring $\{\ObjectPose,\RobotGrasp\}$ from the \ac{SRL}. Fig.~\ref{fig:experiments} shows examples of the PR2 robot handing over objects using our \ac{SRL} model for detection of $\RobotGrasp$ and corresponding final $\ObjectPose$, while considering user defined \mbox{upcoming task and arm mobility capacity}\footnote{Complementary video: \url{https://youtu.be/cFsAEpSn_LI}}.

     \begin{table}[b!]
        \renewcommand{\arraystretch}{1.25}
         \centering
         \begin{tabular}{p{1.1cm}|l|p{1cm}|p{0.7cm}|p{0.8cm}}
         \hline
          \multirow{2}{*}{\textbf{Shape}} &  \multirow{2}{*}{\textbf{Objects}} & \multicolumn{2}{c|}{\textbf{Accuracy}} & \multirow{2}{*}{\textbf{Average}} \\ \cline{3-4}
             & &$\boldsymbol{\ObjectPose}$ &  $\boldsymbol{\RobotGrasp}$ &  \\ \hline

              Cubic & shoe, stapler, camera & 91.4\% & 89.2\% & 90.3\%\\ \hline
              Spherical &  bowl, plate & 93.8\%  & 91.6\% & 92.7\%\\ \hline
              Irregular &  sunglasses, toothbrush & 92.6\%  & 87.7\% & 90.2\% \\ \hline
              Cylindrical & marker, flashlight, glass & 93.1\% & 87.1\% & 90.1\% \\ \hline

         \end{tabular}
         \caption{Handover accuracy performance.\label{tb:grasp}}
     \end{table}

\vspace{-0.25cm}
\section{Conclusions and Future Work}\label{sec:contribution}

    We present a method to reason about suitable object and
    robot grasp configurations for a handover task, subject to the receiver's arm mobility level and the anticipated use of the object. We start by designing a heuristic-guided cost model that adapts handovers to receivers with low arm mobility. Then, through a user study with receivers of different arm mobility capacities, we extract preferences over different handover methods to finally learn them using a \ac{SRL}. Our proposal motivates future research in different directions, including but not limited to (i)~in-person human-robot interaction settings to study acceptance levels on task duration and the naturalness of the robot motion, (ii)~study of failure and recovery alternatives for cases when the robot grasp is not socially acceptable for the handover task, and ways to enrich our \ac{SRL} model to prevent such scenarios and, (iii)~to extend the \ac{SRL} to differentiate non-interactive and interactive tasks, and task dynamics \cite{guevara2017adaptable} - thus, autonomously assisting in home environments. 
    

\vspace{-1em}
\section{Acknowledgements}

    Thanks to the \ac{CHSS} volunteers. Special thanks to Simon Smith and Michael Burke for their peer support.
\vspace{-0.25cm}
\bibliographystyle{ieeetr}
\bibliography{bibliography}

\begin{thebibliography}{10}

\bibitem{ortenzi2020object}
V.~Ortenzi, A.~Cosgun, T.~Pardi, W.~Chan, E.~Croft, and D.~Kulic, ``Object
  handovers: a review for robotics,'' {\em Preprint arXiv:2007.12952}, 2020.

\bibitem{sampson2015using}
P.~Sampson, C.~Freeman, S.~Coote, S.~Demain, P.~Feys, K.~Meadmore, and A.-M.
  Hughes, ``Using functional electrical stimulation mediated by iterative
  learning control and robotics to improve arm movement for people with
  multiple sclerosis,'' {\em IEEE Transactions on Neural Systems and
  Rehabilitation Engineering}, vol.~24, no.~2, pp.~235--248, 2015.

\bibitem{cakmak2011human}
M.~Cakmak, S.~S. Srinivasa, M.~K. Lee, J.~Forlizzi, and S.~Kiesler, ``Human
  preferences for robot-human hand-over configurations,'' in {\em 2011 IEEE/RSJ
  International Conference on Intelligent Robots and Systems}, pp.~1986--1993,
  IEEE, 2011.

\bibitem{sisbot2012human}
E.~A. Sisbot and R.~Alami, ``A human-aware manipulation planner,'' {\em IEEE
  Transactions on Robotics}, vol.~28, no.~5, pp.~1045--1057, 2012.

\bibitem{nemlekar2019object}
H.~Nemlekar, D.~Dutia, and Z.~Li, ``Object transfer point estimation for fluent
  human-robot handovers,'' in {\em 2019 International Conference on Robotics
  and Automation (ICRA)}, pp.~2627--2633, IEEE, 2019.

\bibitem{chan2014determining}
W.~P. Chan, Y.~Kakiuchi, K.~Okada, and M.~Inaba, ``Determining proper grasp
  configurations for handovers through observation of object movement patterns
  and inter-object interactions during usage,'' in {\em 2014 IEEE/RSJ
  International Conference on Intelligent Robots and Systems}, pp.~1355--1360,
  IEEE, 2014.

\bibitem{bestick2016implicitly}
A.~Bestick, R.~Bajcsy, and A.~D. Dragan, ``Implicitly assisting humans to
  choose good grasps in robot to human handovers,'' in {\em International
  Symposium on Experimental Robotics}, pp.~341--354, Springer, 2016.

\bibitem{admoni2014deliberate}
H.~Admoni, A.~Dragan, S.~S. Srinivasa, and B.~Scassellati, ``Deliberate delays
  during robot-to-human handovers improve compliance with gaze communication,''
  in {\em Proceedings of the 2014 ACM/IEEE international conference on
  Human-robot interaction}, pp.~49--56, 2014.

\bibitem{moon2014meet}
A.~Moon, D.~M. Troniak, B.~Gleeson, M.~K. Pan, M.~Zheng, B.~A. Blumer,
  K.~MacLean, and E.~A. Croft, ``Meet me where i'm gazing: how shared attention
  gaze affects human-robot handover timing,'' in {\em Proceedings of the 2014
  ACM/IEEE international conference on Human-robot interaction}, pp.~334--341,
  2014.

\bibitem{shi2013model}
C.~Shi, M.~Shiomi, C.~Smith, T.~Kanda, and H.~Ishiguro, ``A model of
  distributional handing interaction for a mobile robot.,'' in {\em Robotics:
  science and systems}, pp.~24--28, 2013.

\bibitem{huang2015adaptive}
C.-M. Huang, M.~Cakmak, and B.~Mutlu, ``Adaptive coordination strategies for
  human-robot handovers.,'' in {\em Robotics: science and systems}, vol.~11,
  Rome, Italy, 2015.

\bibitem{huber2008human}
M.~Huber, M.~Rickert, A.~Knoll, T.~Brandt, and S.~Glasauer, ``Human-robot
  interaction in handing-over tasks,'' in {\em RO-MAN 2008-The 17th IEEE
  International Symposium on Robot and Human Interactive Communication},
  pp.~107--112, IEEE, 2008.

\bibitem{tang2020assessment}
K.-H. Tang, C.-F. Ho, J.~Mehlich, and S.-T. Chen, ``Assessment of handover
  prediction models in estimation of cycle times for manual assembly tasks in a
  human--robot collaborative environment,'' {\em Applied Sciences}, vol.~10,
  no.~2, p.~556, 2020.

\bibitem{cini2019choice}
F.~Cini, V.~Ortenzi, P.~Corke, and M.~Controzzi, ``On the choice of grasp type
  and location when handing over an object,'' {\em Science Robotics}, vol.~4,
  no.~27, p.~eaau9757, 2019.

\bibitem{pan2017automated}
M.~K. Pan, V.~Skjerv{\o}y, W.~P. Chan, M.~Inaba, and E.~A. Croft, ``Automated
  detection of handovers using kinematic features,'' {\em The International
  Journal of Robotics Research}, vol.~36, no.~5-7, pp.~721--738, 2017.

\bibitem{ortenzi2020grasp}
V.~Ortenzi, F.~Cini, T.~Pardi, N.~Marturi, R.~Stolkin, P.~Corke, and
  M.~Controzzi, ``The grasp strategy of a robot passer influences performance
  and quality of the robot-human object handover,'' {\em Front. Robot. AI 7:
  542406. doi: 10.3389/frobt}, 2020.

\bibitem{aleotti2014affordance}
J.~Aleotti, V.~Micelli, and S.~Caselli, ``An affordance sensitive system for
  robot to human object handover,'' {\em International Journal of Social
  Robotics}, vol.~6, no.~4, pp.~653--666, 2014.

\bibitem{parastegari2017modeling}
S.~Parastegari, B.~Abbasi, E.~Noohi, and M.~Zefran, ``Modeling human reaching
  phase in human-human object handover with application in robot-human
  handover,'' in {\em 2017 IEEE/RSJ International Conference on Intelligent
  Robots and Systems (IROS)}, pp.~3597--3602, IEEE, 2017.

\bibitem{mainprice2012sharing}
J.~Mainprice, M.~Gharbi, T.~Sim{\'e}on, and R.~Alami, ``Sharing effort in
  planning human-robot handover tasks,'' in {\em 2012 IEEE RO-MAN: The 21st
  IEEE International Symposium on Robot and Human Interactive Communication},
  pp.~764--770, IEEE, 2012.

\bibitem{schweighofer2015effort}
N.~Schweighofer, Y.~Xiao, S.~Kim, T.~Yoshioka, J.~Gordon, and R.~Osu, ``Effort,
  success, and nonuse determine arm choice,'' {\em Journal of neurophysiology},
  vol.~114, no.~1, pp.~551--559, 2015.

\bibitem{ardon2019learning}
P.~Ard{\'o}n, {\`E}.~Pairet, R.~P.~A. Petrick, S.~Ramamoorthy, and K.~S. Lohan,
  ``Learning grasp affordance reasoning through semantic relations,'' {\em IEEE
  Robotics and Automation Letters}, vol.~4, no.~4, pp.~4571--4578, 2019.

\bibitem{ardon2020self}
P.~Ard{\'o}n, {\`E}.~Pairet, Y.~Petillot, R.~P.~A. Petrick, S.~Ramamoorthy, and
  K.~S. Lohan, ``Self-assessment of grasp affordance transfer,'' in {\em 2020
  IEEE/RSJ International Conference on Intelligent Robots and Systems (IROS)},
  IEEE, 2020.

\bibitem{choi2009list}
Y.~S. Choi, T.~Deyle, T.~Chen, J.~D. Glass, and C.~C. Kemp, ``A list of
  household objects for robotic retrieval prioritized by people with als,'' in
  {\em 2009 IEEE International Conference on Rehabilitation Robotics},
  pp.~510--517, IEEE, 2009.

\bibitem{richardson2006markov}
M.~Richardson and P.~Domingos, ``Markov logic networks,'' {\em Machine
  learning}, vol.~62, no.~1-2, pp.~107--136, 2006.

\bibitem{jain2009markov}
D.~Jain, P.~Maier, and G.~Wylezich, ``Markov logic as a modelling language for
  weighted constraint satisfaction problems,'' in {\em Eighth International
  Workshop on Constraint Modelling and Reformulation, in conjunction with CP},
  2009.

\bibitem{chen2019pose}
X.~Chen, G.~Wang, H.~Guo, and C.~Zhang, ``Pose guided structured region
  ensemble network for cascaded hand pose estimation,'' {\em Neurocomputing},
  2019.

\bibitem{breitenstein2008real}
M.~D. Breitenstein, D.~Kuettel, T.~Weise, L.~Van~Gool, and H.~Pfister,
  ``Real-time face pose estimation from single range images,'' in {\em 2008
  IEEE Conference on Computer Vision and Pattern Recognition}, pp.~1--8, IEEE,
  2008.

\bibitem{guevara2017adaptable}
T.~L. Guevara, N.~K. Taylor, M.~U. Gutmann, S.~Ramamoorthy, and K.~Subr,
  ``Adaptable pouring: Teaching robots not to spill using fast but approximate
  fluid simulation,'' in {\em Conference on Robot Learning (CoRL)}, 2017.

\end{thebibliography}
\end{document}